\documentclass[10pt,twocolumn,letterpaper]{article}

\usepackage[pagenumbers]{wacv} 

\usepackage{graphicx}
\usepackage{amsmath}
\usepackage{amssymb}
\usepackage{booktabs}

\usepackage[pagebackref,breaklinks,colorlinks]{hyperref}

\usepackage{bm}
\usepackage{multirow}

\usepackage[capitalize]{cleveref}
\crefname{section}{Sec.}{Secs.}
\Crefname{section}{Section}{Sections}
\Crefname{table}{Table}{Tables}
\crefname{table}{Tab.}{Tabs.}

\def\wacvPaperID{2423} 
\def\confName{WACV}
\def\confYear{2025}

\begin{document}

\title{Towards High-fidelity Head Blending with Chroma Keying \\ for Industrial Applications}

\author{Hah Min Lew$^{1}$\footnotemark[1], Sahng-Min Yoo$^{2}$\footnotemark[1] \footnotemark[3], Hyunwoo Kang$^3$\footnotemark[1] \footnotemark[3], and Gyeong-Moon Park$^{4}$\footnotemark[2]\\
$^1$Klleon AI Research, $^2$Samsung Research, $^3$Hyperconnect, $^4$Kyung Hee University\\
{\tt\small hahmin.lew@klleon.io} {\tt\small \{yoosahngmin, khw7147\}@gmail.com}  {\tt\small gmpark@khu.ac.kr}
}

\twocolumn[{
\renewcommand\twocolumn[1][]{#1}
\maketitle
\begin{center}
    \centering
    \captionsetup{type=figure}
\includegraphics[width=2.0782\columnwidth]{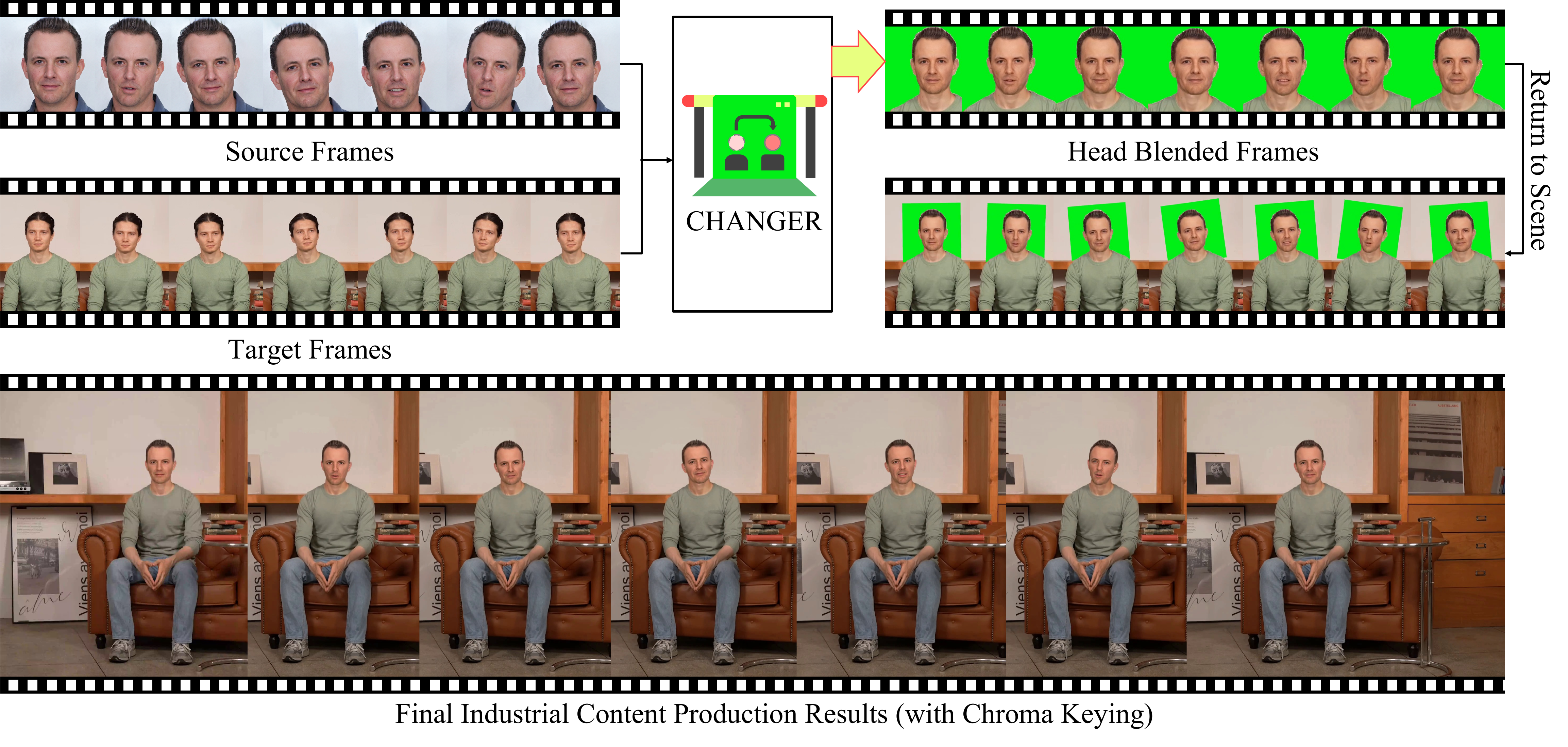}
    \captionof{figure}{\textbf{Illustration of our CHANGER pipeline.} After acquiring the actor's frames (source), we can seamlessly blend acting scenes into the desired scenes with our CHANGER. Chroma keying ensures high-fidelity backgrounds. Here, both of the source and the target actors are virtual humans.
    }
    \label{fig:pipeline}
\end{center}
}]

\begin{abstract} 
    We introduce an industrial Head Blending pipeline for the task of seamlessly integrating an actor’s head onto a target body in digital content creation. The key challenge stems from discrepancies in head shape and hair structure, which lead to unnatural boundaries and blending artifacts. Existing methods treat foreground and background as a single task, resulting in suboptimal blending quality. To address this problem, we propose \textbf{CHANGER}, a novel pipeline that decouples background integration from foreground blending. By utilizing chroma keying for artifact-free background generation and introducing \textit{\textbf{Head shape and long Hair} \textbf{augmentation}} (\textbf{\textit{$\bm{H^2}$ augmentation}}) to simulate a wide range of head shapes and hair styles, CHANGER improves generalization on innumerable various real-world cases.
    Furthermore, our \textit{\textbf{Foreground} \textbf{Predictive} \textbf{Attention} \textbf{Transformer}} (\textbf{\textit{FPAT}}) module enhances foreground blending by predicting and focusing on key head and body regions.
    Quantitative and qualitative evaluations on benchmark datasets demonstrate that our CHANGER outperforms state-of-the-art methods, delivering high-fidelity, industrial-grade results.
\end{abstract} 

{\renewcommand*{\thefootnote}{}%
  \footnotetext{Project page: \href{https://hahminlew.github.io/changer/}{https://hahminlew.github.io/changer}}
{\renewcommand*{\thefootnote}{\fnsymbol{footnote}}\stepcounter{footnote}%
  \footnotetext{Equal contribution}
{\renewcommand*{\thefootnote}{\fnsymbol{footnote}}\stepcounter{footnote}%
  \footnotetext{Corresponding author}
{\renewcommand*{\thefootnote}{\fnsymbol{footnote}}\stepcounter{footnote}%
  \footnotetext{Work done at Klleon AI Research}
\setcounter{footnote}{0}

\begin{figure*}[t]
\begin{center}
\includegraphics[width=2.1\columnwidth]{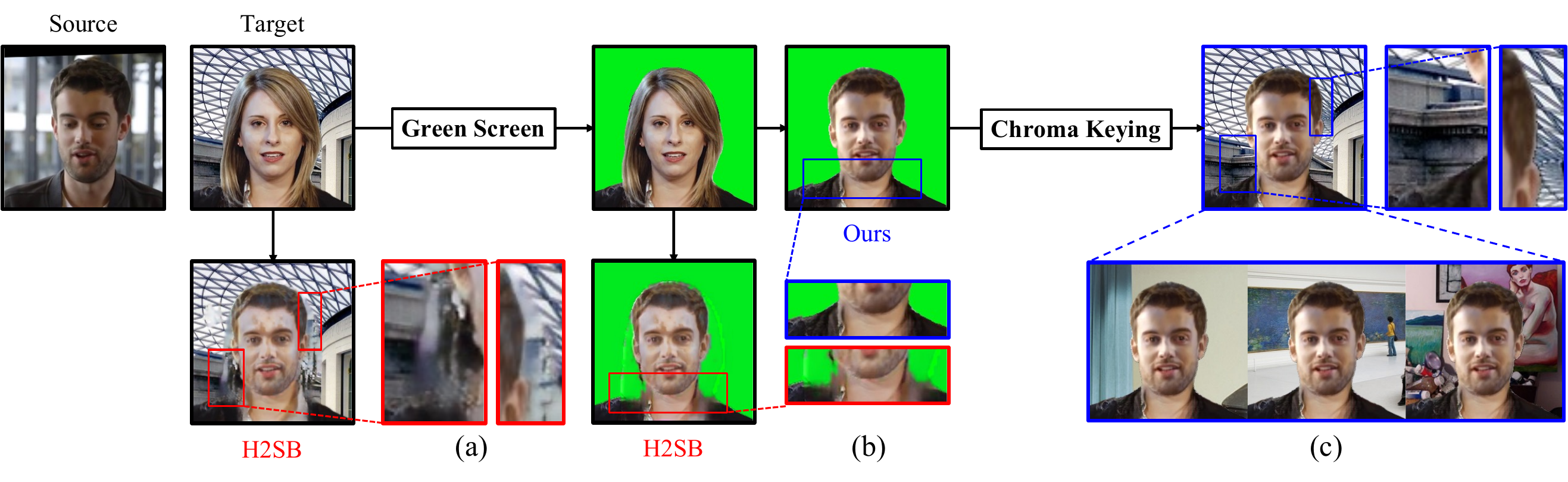}
\end{center}
\vspace{-5mm}
\caption{
\textbf{Motivations of our work}. 
We propose CHANGER to consider the real-world application. As shown in (a), the existing work (H2SB~\cite{shu2022few}) shows severe artifacts on inpainting regions. To inpaint the background flawlessly, we propose to introduce chroma keying in the head blending framework. 
However, it still shows low-fidelity results to inpaint the body, which is hidden due to the head shape and hair difference described in a red box of (b). CHANGER generates the high-fidelity foreground with $H^2$ augmentation and Foreground Predictive Attention Transformer (FPAT), which is explained in Section~\ref{sec:h2_augmentation} and ~\ref{sec:fpat}, respectively. CHANGER removes artifacts as shown in the blue boxes of (b) and (c), and easily changes various high-fidelity real-world backgrounds.
All backgrounds in the figure are from the benchmark dataset~\cite{quattoni2009recognizing}.
}
\label{fig:motivation}
\end{figure*}

\section{Introduction} \label{sec:intro}

In the realm of modern digital content creation, Head Blending, the seamless integration of an actor’s head onto a body filmed in separate takes or contexts is a critical yet under-explored task. We focus on a such process, which is essential for various applications such as visual effects (VFX) post-production, digital human creation, and virtual avatar generation. In these scenarios, integrating an actor’s head with footage where the body and surrounding environment may differ significantly is often necessary. 

The main challenge of Head Blending arises from the discrepancies between the actor's head and the target body, including differences in head shape and hair structure. These discrepancies often lead to unnatural boundaries or blending artifacts, which can be particularly problematic in professional applications where high fidelity and visual coherence are ultimate. The existing method, Head2Scene Blender (H2SB) \cite{shu2022few}, approaches this task by treating the foreground and background generation as a single process. H2SB shows unsatisfactory results (Figure~\ref{fig:motivation}(a), (b)), especially around the boundary regions. Although the generation region has two distinct background and foreground parts, H2SB considers the region at once, which results in an unclear border of a human and artifacts. Moreover, H2SB lacks in mimicking the cross-identity head blending and fails to cover large inpainting regions.

To this end, we propose \textbf{\textit{CHANGER}}, a novel pipeline for \textbf{\textit{C}}onsistent \textbf{\textit{H}}ead blending with predictive \textbf{\textit{A}}tte\textbf{\textit{N}}tion \textbf{\textit{G}}uided foreground \textbf{\textit{E}}stimation under chroma key setting for indust\textbf{\textit{R}}ial applications. We decompose the problem into two distinct sub-tasks: background integration and foreground blending. This decomposition allows for a more focused treatment of each aspect of the task, ensuring higher fidelity in both the background and foreground.

The background integration challenge is addressed by incorporating chroma keying~\cite{shimoda1989new}, a widely used technique in content production where a uniformly colored background (e.g., a green screen) is replaced with the desired scene. This allows for flawless background generation, eliminating the artifacts that arise when the foreground and background are blended simultaneously. By decoupling the foreground blending from the background, we ensure that the visual integrity of the scene is preserved, even in complex environments.

For the foreground blending, we tackle the problem of seamlessly integrating the actor’s head onto the body of the target, particularly in cases where significant differences exist in head shape and hair structure. To generate the high-fidelity foreground, we devise two contributions, one from a data-centric and the other from a model-centric perspective. 

First, we propose a novel data augmentation technique called \textit{\textbf{H}ead shape and long \textbf{H}air \textbf{augmentation}} (\textbf{\textit{$\bm{H^2}$ augmentation}}), which simulates a wide range of head shapes and hair styles in the self-supervised training. This enables our model to better generalize to real-world variations and handle the significant visual discrepancies that often arise in professional content production.

Second, we introduce the \textbf{\textit{F}}oreground \textbf{\textit{P}}redictive \textbf{\textit{A}}ttention \textbf{\textit{T}}ransformer (\textbf{\textit{FPAT}}), a novel architecture designed for foreground blending. FPAT predicts the exact regions of the head and body that require attention and apply targeted attention to these areas during the blending process. By explicitly restricting the attention to these key regions, FPAT enhances the blending quality, particularly in areas where head shape and hair differences pose a challenge.

To summarize, we propose the first comprehensive solution for the Head Blending task in industrial content production. Unlike the previous approach that treats this process as part of general head or face swapping, our method focuses explicitly on the seamless blending of an actor’s head with the target body, ensuring realistic and high-quality results. Our method, CHANGER, significantly outperforms state-of-the-art techniques, as demonstrated through both quantitative metrics and qualitative evaluations on benchmark datasets.

\begin{figure*}[t]
\begin{center}
\includegraphics[width=2.1\columnwidth]{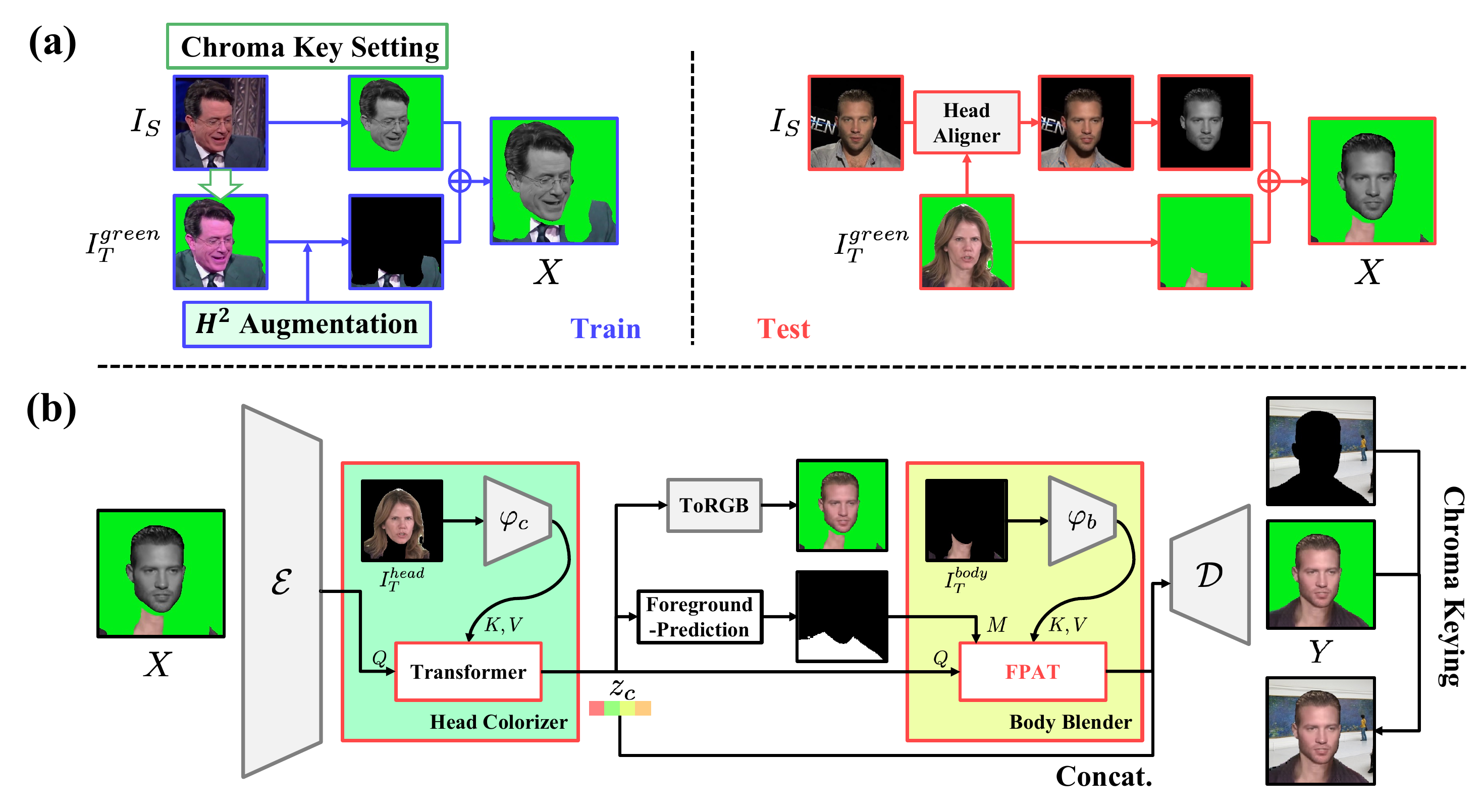}
\end{center}
\vspace{-5mm}
\caption{
\textbf{Network overview of CHANGER.} 
(a) We visualize how we conduct the input of the network ($X$) at the train (blue) and the test (red). We apply $H^2$ augmentation during the training to improve the fidelity of the generated image by improving the diversity of the input. (b) We visualize the network of CHANGER. The head colorizer colorizes the gray head of $X$, and the body blender inpaints the hidden body with a foreground mask-aware attention mechanism. Please refer to the detailed explanations of $H^2$ augmentation and FPAT in Section~\ref{sec:h2_augmentation} and~\ref{sec:fpat}, respectively.
 }
\label{fig:network}
\end{figure*}


In summary, our main contributions are as follows:
\begin{itemize}
    \item We propose CHANGER, a novel pipeline that utilizes chroma keying for the first time to decouple background integration from the head blending process, addressing the common artifacts seen in prior methods.
    \item We introduce $H^2$ augmentation, a data-centric approach designed to handle significant variations in head shape and hair structure, enhancing the robustness of the Head Blending process.
    \item We present the FPAT module, which uses predictive attention to focus on key regions of the head and body, resulting in high-fidelity blending with minimized artifacts.
    \item CHANGER significantly outperforms existing methods on benchmark datasets, both quantitatively and qualitatively, showcasing its effectiveness in industrial content production scenarios.
\end{itemize}

\section{Related Work}



\noindent\textbf{Head Blending.} \
Head Blending aims to replace the head in a target image with a source head, ensuring that the result is seamless and maintains the consistency of the skin color of the target image.
To address this problem, H2SB~\cite{shu2022few} proposes a semantic-guided color reference creation module based on~\cite{zhang2020cross} for re-coloring the head and filling the neck and the background at once.
However, we empirically find that H2SB results in inadequate generation results which is unsuitable for real-world application.
H2SB~\cite{shu2022few} relies on a single feature correspondence matrix, which is proposed for image translation~\cite{zhang2020cross} and has a simple U-Net structure.
In contrast, we propose a novel approach that manages the background with chroma keying for high-quality and efficient background changes.

\noindent\textbf{Mask-Aware Transformers.} \ Transformer~\cite{vaswani2017attention} is a model that processes a sequence of tokens with an attention mechanism. The original attention in the transformer computes the similarity between the query and the key, where all tokens have participated. On the other hand, there are lines of work that restrict the region of computing attention using the pre-defined masks that reflect the prior knowledge of the task. Transformer decoder block~\cite{vaswani2017attention} uses causal attention in a transformer decoder to block that later tokens affect the previous tokens to be generated. Swin Transformers~\cite{liu2021swin} and ConViT~\cite{d2021convit} restrict or prioritize the attention region in spatially closed patches to inject the spatial inductive bias to the model, and OAMixer~\cite{kang2022oamixer} reweighs the attention with a soft mask to strengthen the relationship between semantically related tokens to improve generalization and mitigates with background bias.
In this work, we design a novel foreground predictive attention transformer (FPAT) by leveraging a neck and body region prediction module. FPAT differs from previous mask-aware transformers in that the mask for attention is not given, but we predict the foreground mask implicitly within training our head blending pipeline.     
\section{Method}

In this section, we present the detailed methodology behind CHANGER, our novel framework for Head Blending. CHANGER addresses the primary challenges of blending an actor’s (source) head onto a target body by decomposing the task into two key sub-tasks: background integration and foreground blending.
CHANGER handles the background integration via combining chroma keying. We detail the network input and output preparation for chroma keying in Section~\ref{sec:chromakey}. To generate high-fidelity foregrounds, we propose new data augmentation and model design. To address the constraints of self-supervised training that uses the same images for both source and target, where cross-identity settings lack ground truth, we propose a novel augmentation method called $H^2$ augmentation. This technique broadens the diversity of the input $X$, enhancing the ability of the model to adapt to various identities. Further details on this approach can be found in Section~\ref{sec:h2_augmentation}. In Section~\ref{sec:fpat}, we detail our Foreground Predictive Attention Transformer (FPAT) which enhances foreground blending.
The overall network of CHANGER is shown in Figure~\ref{fig:network}. 

\subsection{Chroma Keying for Head Blending}
\label{sec:chromakey}
We propose a chroma key setting for a head blending task to divide the labor of generating the background region to chroma keying. To this end, we modify the input of the network $X$ to have a green background, ensuring the output of the network $Y$ also maintains this green background as shown in Figure~\ref{fig:network}(a). To do so, we first paint the background of the target image $I_T$ as green and acquire $I_T^{\text{green}}$ by finding the foreground with the state-of-the-art face parsing network \cite{yu2018bisenet}. Then, we extract the head mask of the source, $\mathtt{M}^{head}_{S}$,  and the target head mask $\mathtt{M}^{head}_{T}$. 
We obtain a union mask of the source head mask $\mathtt{M}^{head}_{S}$ and the target head mask $\mathtt{M}^{head}_{T}$ as follows:
\begin{equation}
   \label{eq:1}
   \mathtt{M}^{head}_{union} =
   \begin{cases}
       \mathtt{M}^{head}_{S} \oplus \mathtt{M}^{head}_{h^2}, & \text{(train)},\\
       \mathtt{M}^{head}_{S} \oplus \mathtt{M}^{head}_{T}, & \text{(test)},
   \end{cases}
\end{equation}
where $\oplus$ is a union operation. Note that we use $\mathtt{M}^{head}_{h^2}$, which is the output of $H^2$ augmentation during training to obtain $\mathtt{M}^{head}_{union}$ since the ${M}^{head}_{S}$ and ${M}^{head}_{T}$ is identical in the self-supervised setting. Finally, we formulate input $X$ as the following equation:
\begin{equation}
\label{eq:3}
    X = I^{gray}_{S} + I_T^{green} \otimes (1 - \mathtt{M}^{head}_{union}) + I^{green} \otimes \mathtt{M}^{ip},
\end{equation}
where the gray-scale source head, $I_S^{gray} = \mathtt{g}(I_S \otimes \mathtt{M}^{head}_{S})$, where $\mathtt{g}(x)$ indicates a gray-scaling function and $\otimes$ is the pixel-wise multiplication, and an inpainting mask $\mathtt{M}^{ip} = \mathtt{M}^{head}_{union} - \mathtt{M}^{head}_{S}$. 
During test environments, any targets are acceptable in our pipeline.
If the target is filmed on a green screen, we directly apply our CHANGER.

\subsection{\boldmath $H^2$ Augmentation}
\label{sec:h2_augmentation}

Since we train the model in a self-driven manner during training, the target image is generated from the source image. We propose a simple but powerful $H^2$ augmentation that manipulates the input $X$ during the self-identity head blending training to simulate the various cross-identity blending scenarios, especially the settings where the foreground blending region is large. 
Existing methods lack variation in inpainting regions, critical for self-driven training. 
To tackle the issue, we carefully designed promising computer vision techniques and the stochastic sampling method.
Since the head shape and the hair difference between the source and the target generates a large mismatch region, $H^2$ augmentation includes a head shape and a long hair augmentation.

\begin{figure*}[t]
\begin{center}
\includegraphics[width=2.1\columnwidth]{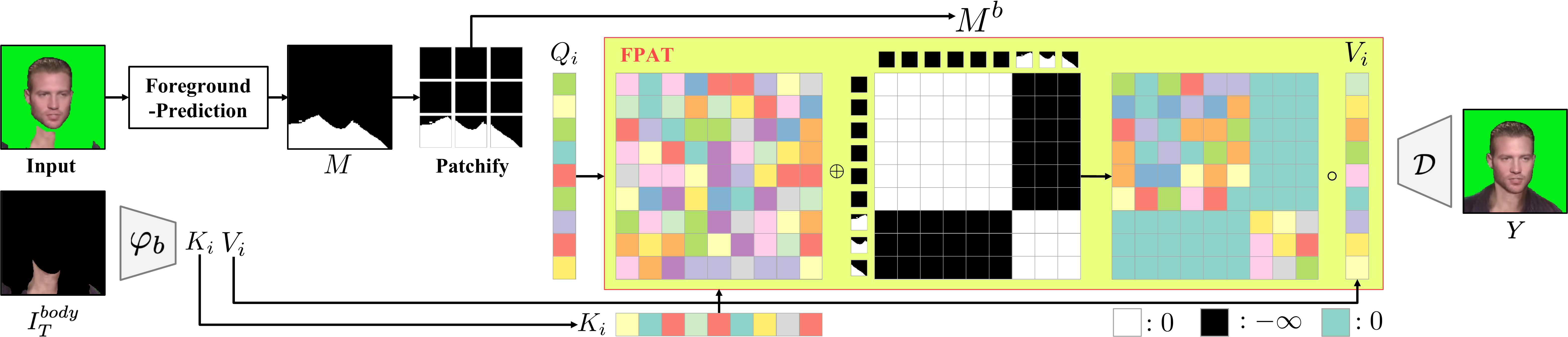}
\end{center}
\vspace{-5mm}
\caption{
\textbf{Visualization of attention mechanism of our Foreground Predictive Attention Transformer (FPAT) block.} The Foreground-Prediction module predicts the foreground mask $M$ of the body and the neck region, and the attention is reweighted according to $M$.
}
\label{fig:fpat}
\end{figure*}

\noindent\textbf{Head Shape Augmentation.} Since an outline of a source head differs from a target head in the cross-identity blending scenarios, an empty region between a source head and a target body is quite diverse.
To mimic possible mismatches appearing in the blending procedure under a self-supervised manner, we randomly augment the region by transforming the source head mask with a head shape augmentation $\mathcal{T}_{head}$, which includes an affine transformation, squeezing, expanding, and varying dilation widths. Therefore, from the source head mask $\mathtt{M}_S^{head}$, we obtain the augmented head mask as following:
\begin{equation}
\label{eq:head}
\mathtt{M}_{h^{1}}^{head} = \mathcal{T}_{head}(\mathtt{M}_S^{head}).
\end{equation}

\noindent\textbf{Long Hair Augmentation.} To mimic the hair differences in the cross-identity setting, we randomly sample an identity whose hair is long enough to cover its clothing and body. With the hair mask of the sampled identities $\mathtt{M}_{long}^{hair}$, we apply the long hair augmentation $\mathcal{T}_{hair}$ to the augmented head mask $\mathtt{M}_{h^{1}}^{head}$, and get the final head mask as follows:
\begin{equation}
    \label{eq:long_hair}
    \mathtt{M}^{head}_{h^{2}}= \mathcal{T}_{hair}(\mathtt{M}_{h^{1}}^{head}) =
    \begin{cases}
        \mathtt{M}^{head}_{h^{1}} \oplus \mathtt{M}^{hair}_{long},& \text{if } p\geq \epsilon,\\
        \mathtt{M}^{head}_{h^{1}}, & \text{otherwise},
    \end{cases}
\end{equation}
where $p$ is sampled from uniform distribution and $\epsilon \in (0, 1)$ is a fixed threshold. 
The visualizations of Eq. (\ref{eq:head}) and (\ref{eq:long_hair}) are shown in our supplementary material.
With the augmented head mask $\mathtt{M}^{head}_{h^{2}}$, we obtain the union mask with the source head mask $\mathtt{M}^{head}_{union} = \mathtt{M}^{head}_{h^{2}} \oplus \mathtt{M}^{head}_{S}$ then produce the input $X$ with Eq. (\ref{eq:3}).

Our proposed $H^2$ augmentation creates various union masks from the source head and enables diverse neck and body completion regions in a self-supervised manner.  $H^2$ augmentation is a delicate solution that addresses unique characteristics during head blending training. 
The ablation study in Table~\ref{tab:ablation_h2aug} and Figure~\ref{fig:ablation} demonstrates the novelty of $H^2$ augmentation. 


\begin{table*}
\small
\centering
\begin{tabular}{ccccc|ccc}
\hline
\textbf{Method} & PSNR $\uparrow$ & LPIPS $\downarrow$ & $L_1$ $\downarrow$ & SSIM $\uparrow$ & FPS $\uparrow$ & MACs $\downarrow$& Param. $\downarrow$ \\ 
\hline
H2SB~\cite{shu2022few} & 12.397 & 0.134 & 0.125 & 0.743 & \multirow{2}{*}{28.10} & \multirow{2}{*}{122.07G} & \multirow{2}{*}{24.37M}\\
H2SB~\cite{shu2022few} + CK & 12.974 & 0.086 & 0.119 & 0.737 & & &\\
\hline
{\textbf{Ours}}
 & \textbf{27.845} & \textbf{0.011} & \textbf{0.014}
 & \textbf{0.950}  & \textbf{60.57} & \textbf{81.73G} & \textbf{8.92M}\\ 
\hline
\end{tabular}
\caption{\textbf{Quantitative comparison with H2SB~\cite{shu2022few} and our CHANGER.} ``CK'' is an abbreviation of chroma keying.
}
\label{tab:quantitative}
\end{table*}
\begin{table}
\setlength{\tabcolsep}{2pt}
\small
\centering
\begin{tabular}{ccccc}
\toprule
\textbf{Method} & \textbf{BG} $\uparrow$ & \textbf{ID} $\uparrow$ & \textbf{Natural} $\uparrow$  & \textbf{Holistic} $\uparrow$\\ \midrule
H2SB~\cite{shu2022few} 
&  0.696&  1.234&  1.035& 0.725\\

H2SB~\cite{shu2022few} + CK & 0.720& 1.188&  0.847&   0.642\\ \midrule
\textbf{Ours} 
& \textbf{1.110} & \textbf{1.300} & \textbf{1.226}  & \textbf{1.091}\\ \bottomrule
\end{tabular}
\caption{\textbf{Quantitative comparison from the user study.} ``CK'' is an abbreviation of chroma keying.
}
\label{tab:userstudy}
\end{table}

\subsection{Foreground Predictive Attention Transformer}
\label{sec:fpat}

The network architecture for the foreground blending is divided into two components: (1) Head Colorizer that transfers the color of the target to the gray-scaled source head, and (2) Body Blender that generates the body for a seamless connection between the source head and the target body via our novel Foreground Predictive Attention Transformer (FPAT).

Head Colorizer in Figure~\ref{fig:network}(b) transfers the color from the head of the target image into the head of the source image. Since the input of the network $X$ has a gray head, the model should colorize the head of the input by referring to the conditioned target image. Head colorizer is composed of cross-attention transformer blocks with a query from the embedded features of the input $X$ using the encoder $\mathcal{E}$ and the key and the value from the target head $I_T^{head}$ using $\varphi_c$, which is a conditional projection embedder~\cite{vaswani2017attention, dosovitskiy2020image, dhariwal2021diffusion, rombach2022high}.
The head colorizer outputs the intermediate hidden representation $ z_c \in \mathbb{R}^{C \times h \times w}$, where $C$ is the number of channels and $(h, w)$ is the resolution of the feature. 

The Body Blender in Figure~\ref{fig:network}(b) generates the occluded body region by incorporating our novel FPAT.
Here, the body blender aims to ensure coherent edge continuation and seamless head-body connection, while avoiding inappropriate influences from background or mismatched regions.
Therefore, the body blender requires a sophisticated attention mechanism that computes region-selective attention: for occluded clothing, exclusively from other clothing regions; for the head-body junction, solely from the facial area. To address these distinctive requirements, we introduce the Foreground Predictive Attention Transformer (FPAT), a novel architecture that redefines masked attention in the context of head blending.

FPAT computes the masked attention between tokens from the intermediate feature of the colorized head ($z_c$) to the feature of the target body $I_T^{body}$.
Here, the masked attention is applied within respective regions, i.e., foreground to foreground and background to background, respectively.
FPAT starts with the output of the head colorizer $z_c$, and predicts a foreground region as a binary mask $ M \in \mathbb{R}^{h \times w} $, as shown in Figure~\ref{fig:fpat}.  
Then, FPAT patchifies the mask $M$ with $N$ patches; $M_i$ for $i \in \{ 1, ..., N \}$. 

With the patchified mask, FPAT computes the binary attention mask $M^b$ as following:
\begin{equation}
    M_{ij}^b=\begin{cases}
    0, & \text{if } M_i \text{ and } M_j \text{ are the same type of patches},\\
    -\infty, & \text{otherwise},
  \end{cases}
\end{equation}
where `the same type of patches' refers to pairs of patches that are either both classified as foreground or both as background, and $M_{ij}^b$ is the $(i, j)$-th element of $M^b$.

FPAT masks the attention between a query from the latent representation $z_c^p$, key and value from the target head feature $z_{body}^p$ as follows:
\begin{equation}
    \text{FPAT}(Q_i,\ K_i,\ V_i) = \text{softmax} (\frac{{Q_i}{K_i^T}}{\sqrt{d_k}} + M^b) \cdot V_i,
\end{equation}
where $Q_i = z_c^p \cdot W^Q_i,\ K_i = z_{body}^p \cdot W^K_i,$ and $V_i = z_{body}^p \cdot W^V_i$. The dimension of learnable projection parameter matrices~\cite{vaswani2017attention} is $W^Q_i \in \mathbb{R}^{N \times D_K},\ W^K_i  \in \mathbb{R}^{N \times D_K},$ and $W^V_i  \in \mathbb{R}^{N \times D_V}$, respectively.

Note that FPAT updates the body and the neck parts of the hidden representation $z_c^p$ by only referring to the body and the neck parts of $z_{body}^p$. Our proposed FPAT \textcolor{black}{generates high-fidelity foregrounds by restricting the attention region with the predicted regions}.

\subsection{Training Objectives}
\label{sec:objective}
We formulate $z$ by channel-wisely concatenating the output of the head colorizer $z_c$ and the output of the body blender $z_b$. $z$ is the input for the decoder $\mathcal{D}$:
\begin{equation}
    \hat{Y} = \mathcal{D}(z),
\end{equation}
where $\hat{Y}$ contains spatial information of the colorized head from $I^{gray}_{S}$ and the the body completion.

We define the final output $Y$ by utilizing $I_{T}$ as follows:
\begin{equation}
    Y = \hat{Y} \otimes \mathtt{M}^{head}_{union} + I_T \otimes (1 - \mathtt{M}^{head}_{union}).
\end{equation}
We use five loss functions, \textbf{(1)} $\mathcal{L}_{rec}= ||Y \otimes \mathtt{M}^{head}_{S} - I_T \otimes \mathtt{M}^{head}_{S}||_1 $, the reconstruction loss for the final output head and the ground truth, \textbf{(2)} $\mathcal{L}_{hc} = ||Y - I^{hc}_{T}||_1$, the reconstruction loss for the output of $\texttt{ToRGB}$ block, \textbf{(3)} $\mathcal{L}_{mask}=|| M_{gt}- M||_1$~\cite{milletari2016v}, the loss for the output of the Foreground-Prediction module where $M_{gt}$ is the ground-truth full body mask, \textbf{(4)} perceptual loss $\mathcal{L}_{per}= \sum^{L}_{i=1}||\Phi_i(Y) - \Phi_i(I_T)||_1$, and \textbf{(5)} adversarial loss $\mathcal{L}_{adv}$.
$I^{hc}_{T}$ is a target image without neck, and body completion and $\Phi$ is a pre-trained VGG19 network~\cite{simonyan2014very}.

The final objective function is as follows:
\begin{align}\label{eq:12}
    \mathcal{L}_{total} = \lambda_{rec} \mathcal{L}_{rec} + \lambda_{hc}\mathcal{L}_{hc} + \lambda_{mask} \mathcal{L}_{mask} \nonumber\\ + \lambda_{per} \mathcal{L}_{per} + \lambda_{adv} \mathcal{L}_{adv},
\end{align}
where   $\lambda_{rec}$, $\lambda_{hc}$, $\lambda_{mask}$, $\lambda_{per}$,   and $\lambda_{adv}$ are weights for the loss $\mathcal{L}_{rec}$, $\mathcal{L}_{hc}$, $\mathcal{L}_{mask}$, $\mathcal{L}_{per}$, and $\mathcal{L}_{adv}$, respectively.

\section{Experiments}





\noindent\textbf{Implementation Details.} \
We combined three different benchmark datasets with training and testing our CHANGER and the state-of-the-art model H2SB~\cite{shu2022few}:
\textbf{(1)} VoxCeleb1~\cite{Nagrani17}, \textbf{(2)} VoxCeleb2~\cite{Chung18b}, and \textbf{(3)} HDTF~\cite{zhang2021flow}.
An Adam optimizer~\cite{kingma2014adam} with hyperparameters of $\beta_{1}=0.9$ and $\beta_{2}=0.999$ was used for every models.
We usedlearning rates 1e-4 and 4e-4 for the generator and discriminator, respectively.
\textcolor{black}{We used $\epsilon=0.5$ in Eq. (\ref{eq:long_hair})}.
We used 4 NVIDIA RTX 3090 (24 GB) GPUs to train our CHANGER. Due to the limitations of GPU resources, the experiments were conducted on 256-resolution images. However, the core components of CHANGER, $H^2$ augmentation and FPAT, are inherently resolution-agnostic. This design ensures that our CHANGER can seamlessly scale to higher resolutions for future applications.

\begin{figure*}[t]
\begin{center}
\includegraphics[width=2.1\columnwidth]{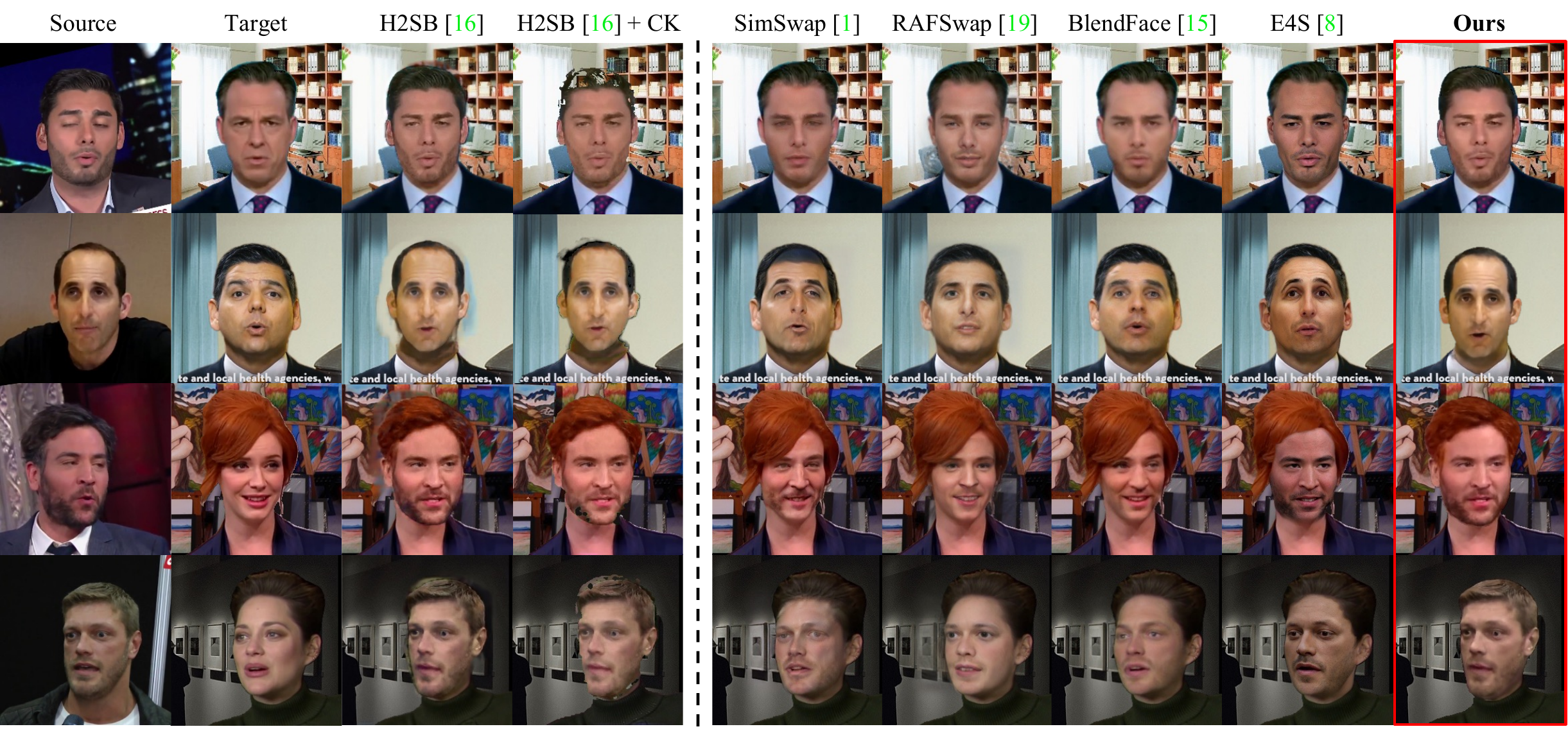}
\end{center}
\vspace{-5mm}
\caption{
We compared qualitative results between CHANGER with the state-of-the-art head blending and face swapping models: H2SB~\cite{shu2022few}, SimSwap~\cite{chen2020simswap}, RAFSwap~\cite{xu2022region}, BlendFace~\cite{shiohara2023blendface}, and E4S~\cite{liu2023fine}.
``CK'' represents inference on the chromakey configuration.
}
\label{fig:qualitative}
\end{figure*}

\subsection{Comparison with the State-of-the-Art}
\label{sec:comparison}

\noindent\textbf{Quantitative Comparison.} \
In order to provide a clear benchmark for advancement in the specific domain of head blending, we quantitatively compared our method with H2SB~\cite{shu2022few}, the only existing method designed for the head blending task. Moreover, we generated the self-blending results and prepared a ground truth to evaluate the performance quantitatively.

Table~\ref{tab:quantitative} shows that our CHANGER outperforms both H2SB and H2SB $+ \ CK$ quantitatively.
In addition, we analyzed the computational efficiency of our method and the baseline method in terms of inference time. Our method achieved comparable speed performance (2.2 times faster \textit{FPS} than H2SB) to the baseline methods while requiring only a fraction of the computational cost (33\% less \textit{MACs} than H2SB) and 64\% fewer parameters (\textit{Param.}). These findings demonstrate the practical feasibility of our method for real-world applications.

To assess perceptual quality of the results, we conducted a user study referring to the state-of-the-art network~\cite{shu2022few}, which rates (1) the fidelity of background regions (\textit{BG}), (2) the identity similarity according to the head skin colorization fidelity (\textit{ID}), (3) the naturalness of generated neck and body (\textit{Natural}), and (4) overall perceptual qualities of head blending images (\textit{Holistic}).
We attached the user study material in supplementary.

The user study results from 21 human evaluators rating of 0 to 2 for various criteria are shown in Table~\ref{tab:userstudy}.
In the blind test between our CHANGER with H2SB and H2SB $+ \ CK$, most of the users ranked our head blending results highest as depicted in the table, even though duplication pick was allowed. 
Ours outperformed the baselines regarding identity preservation and the fidelity of the head and background region with overall user satisfaction.

\noindent\textbf{Qualitative Comparison.} \
Qualitative comparisons were conducted more extensively, encompassing a wider range of comparative face swapping methods.

Figure~\ref{fig:qualitative} shows the head blending results of various source and target images against the state-of-the-art head blending and face swapping frameworks.
As shown in Figure~\ref{fig:qualitative}, our CHANGER shows high-fidelity head blending results with high-quality backgrounds. H2SB shows poor background results. Although H2SB with our chroma keying (H2SB $+ \ CK$) shows slightly improved background outcomes, it still suffers from low-fidelity results on body blending. The body blending results are improved significantly in Ours due to the proposed $H^2$ augmentation and FPAT.


\subsection{Ablation Study}
In our ablation study, we examined the respective efficacy of the CHANGER components by removing or altering the key components.
Table~\ref{tab:ablation_main} and Table~\ref{tab:ablation_h2aug} show the performance of the possible variants of CHANGER based on our main proposal and the details on $H^2$ augmentation, respectively.

\noindent\textcolor{black}{\textbf{Quantitative Results.}} We explored the effectiveness of our key components by removing $H^2$ augmentation (model ``A'') and FPAT (model ``B'') as shown in Table~\ref{tab:ablation_main}.
We replaced FPAT with a conventional cross-attention transformer for model ``B''.
Quantitative results show that our full model achieves the best performance in overall metrics.

We evaluated the components in our proposed $H^2$ augmentation: (1) the head shape augmentation (``Head'') and (2) the long hair augmentation (``Hair'') in Table~\ref{tab:ablation_h2aug}.
Model ``C'' (without the long hair augmentation) showed better performance than model ``D'' (without the head shape augmentation) in terms of SSIM and PSNR.
For LPIPS, model ``D'' shows better results.
These results imply that head shape augmentation leads the better accuracy in the reconstruction of the image, while long hair augmentation allows the better perceptual quality.

\begin{table}[t]
\small
\centering
\setlength{\tabcolsep}{4pt}
\begin{tabular}{c|cc|cccc}
\hline
\textbf{Type} & \textbf{FPAT}  & $\bm{H^2}$ & PSNR $\uparrow$ & LPIPS $\downarrow$ & $L_1$ $\downarrow$ & SSIM $\uparrow$ \\ \hline
A & \checkmark & & 16.965 & 0.122 & 0.067 & 0.863 \\
B & & \checkmark & 27.199 & 0.012 & 0.015 & 0.949 \\
\hline
\textbf{Ours} & \checkmark & \checkmark & \textbf{27.845} & \textbf{0.011} & \textbf{0.014} & \textbf{0.950}    \\  \hline
\end{tabular}
\caption{\textbf{Ablation study on our proposal.} Quantitative results when excluding the proposed $H^2$ augmentation and FPAT, respectively.}
\label{tab:ablation_main}
\end{table}
\begin{table}[t]
\small
\centering 
\setlength{\tabcolsep}{4pt}
\begin{tabular}{c|cc|cccc}
\hline
    \textbf{Type} & \textbf{Head} & \textbf{Hair} & PSNR $\uparrow$ & LPIPS $\downarrow$ & $L_1$ $\downarrow$ & SSIM $\uparrow$  \\ \hline
C & \checkmark & 	& 26.437	& 0.023	& 0.020 & 0.948\\
D & & \checkmark 	& 25.037	& 0.018	& 0.020 & 0.932\\
\hline
\textbf{Ours} & \checkmark & \checkmark  & \textbf{27.845} & \textbf{0.011} & \textbf{0.014} & \textbf{0.950} \\
\hline
\end{tabular}
\caption{\textbf{Ablation study on $\bm{H^2}$ augmentation.} All combinations made by the two components of $H^2$ augmentation were re-trained and evaluated. ``Head'' is the head shape augmentation. ``Hair'' is the long hair augmentation.}
\label{tab:ablation_h2aug}
\end{table}

\begin{figure}[t]
\begin{center}
\includegraphics[width=\columnwidth]{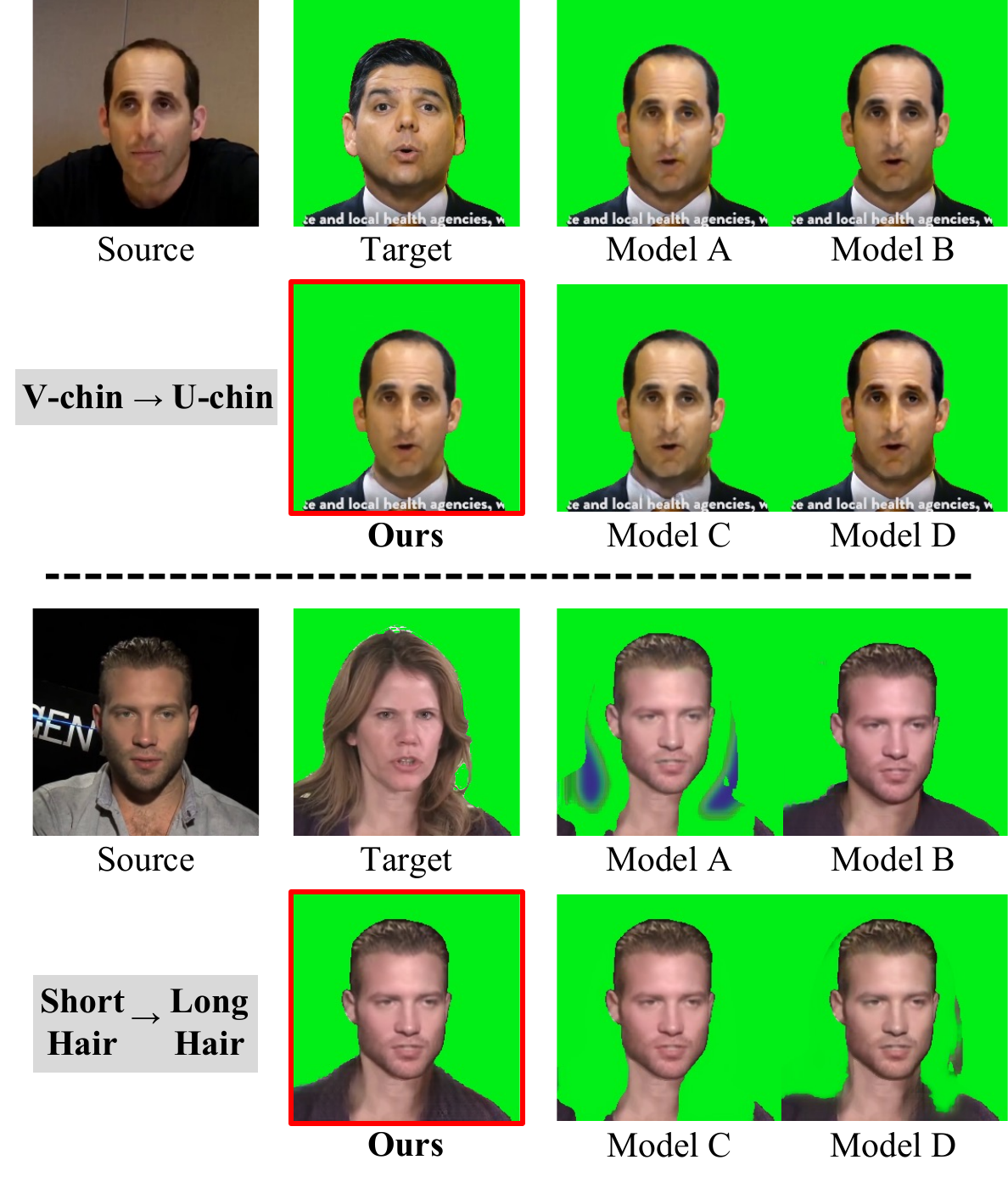}
\end{center}
\vspace{-5mm}
\caption{
\textbf{Ablation study on model designs.}  
Both the head shape augmentation and the long hair augmentation are crucial components for high-fidelity head blended images.
FPAT allows fine-grained foreground generation.
}
\label{fig:ablation}
\end{figure}

\noindent\textcolor{black}{\textbf{Qualitative Results.}} We also qualitatively analyzed the effect of each component by conducting cross-head blending to examine the large differences in head shape and hair, i.e., v-chin to u-chin and short-hair to long-hair as depicted in Figure~\ref{fig:ablation}.
The head shape augmentation allowed the generation of a more acceptable neckline (see top results of models ``B'' and ``C'' compared to models ``A'' and ``D'').
On the other hand, the long-hair augmentation tended to the better generation of hidden regions due to the long hair of the target (see bottom results of models ``B'' and ours compared to models ``A'' and ``C'').
Without $H^2$ augmentation or employing either one, both head blending quality and fidelity deteriorate (see models ``A'', ``C'', and ``D'').
With FPAT, the model showed better foreground blending compared to ours and model ``B''.
The proposed $H^2$ augmentation and FPAT contributed significantly to performing head blending in the chroma key setting. 

\begin{figure}[t]
\begin{center}
\includegraphics[width=\columnwidth]{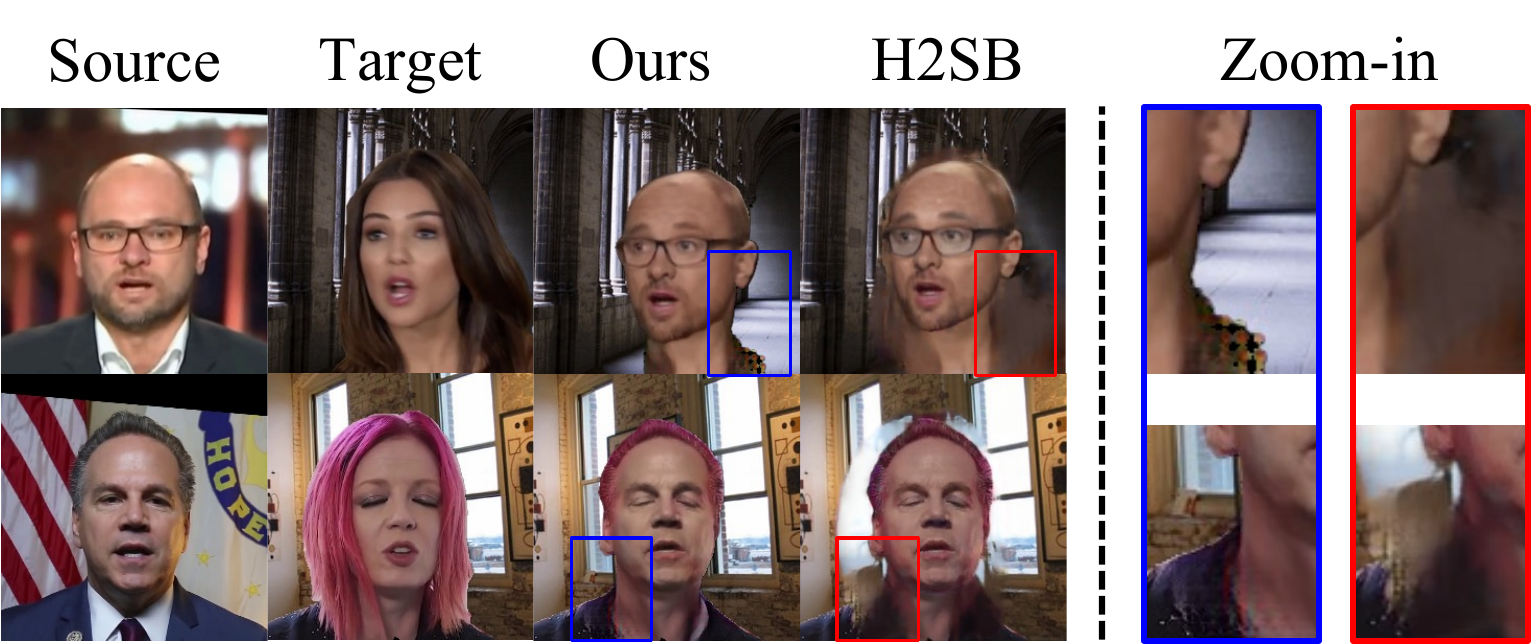}
\end{center}
\vspace{-5mm}
\caption{
\textbf{Limitations.}  
When the target has certain extreme attributes such as too rich hair or pinkish hair, CHANGER suffers artifacts on generated body regions.
}
\label{fig:limitation}
\end{figure}
\begin{figure}[t]
\begin{center}
\includegraphics[width=\columnwidth]{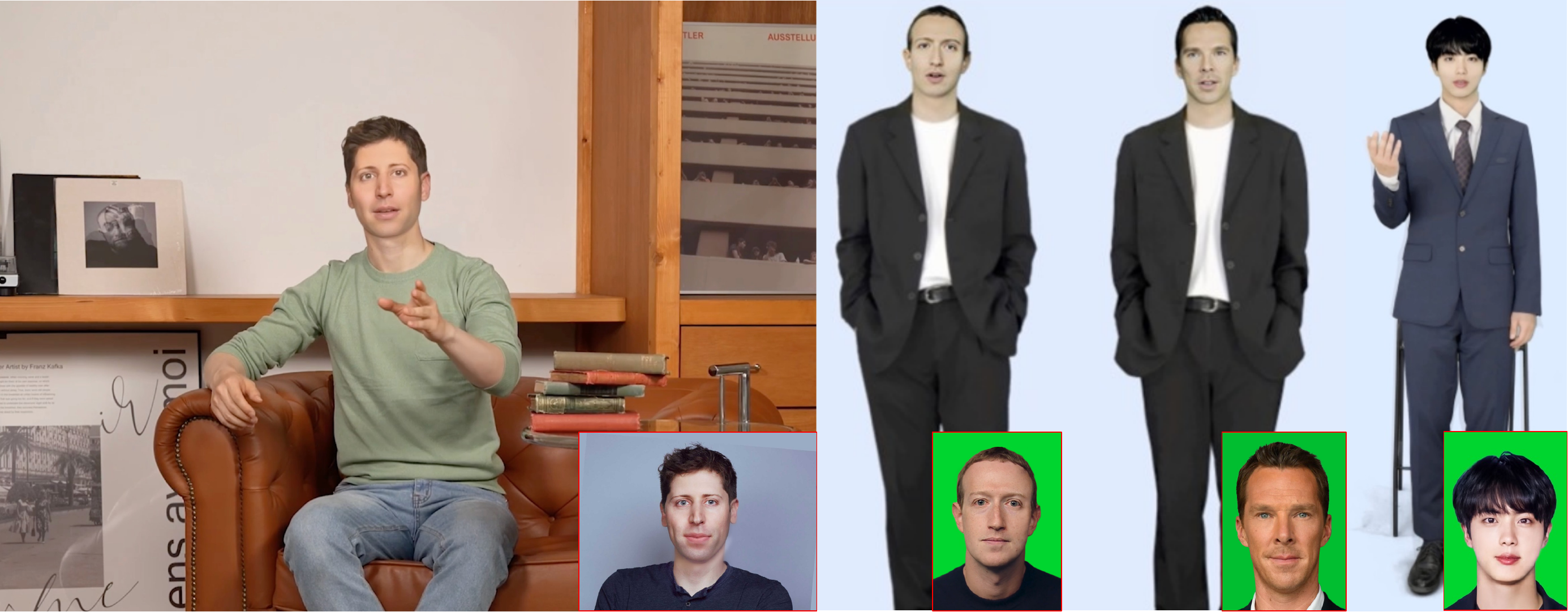}
\end{center}
\vspace{-5mm}
\caption{
\textbf{Various industrial application examples.}  
By leveraging chroma key technique with our proposed CHANGER pipeline, we can obtain various high-fidelity head blended videos in the wild environments. 
The red boxes represent the source images.
}
\label{fig:industrial}
\end{figure}

\subsection{Discussions}

\noindent\textbf{Limitations.} 
As shown in Figure~\ref{fig:limitation}, CHANGER encounters challenges under certain extreme conditions where the target image has too rich hair so that the body region is largely hidden. Also, CHANGER sometimes fails to generate high-fidelity head blending results when the target has extreme attributes, e.g., pinkish hair. Despite these imperfections, we emphasize that across all rows in Figure~\ref{fig:qualitative}, the results of CHANGER consistently surpass the achievements of H2SB, the state-of-the-art of head blending.

\noindent\textbf{Social Impacts.} As shown in Figure~\ref{fig:industrial}, our CHANGER pipeline can achieve various high-fidelity industrial content productions in the wild. 
However, due to its high performance, our technology might cause cultural, political, and ethical social problems, such as indistinguishable deep fake videos, invasion of privacy, or even defamation.
    
\section{Conclusion}
In this work, we presented CHANGER, a novel head blending pipeline for high-fidelity industrial content production within chroma key settings for the first time. Our approach, proposing $H^2$ Augmentation and Foreground Predictive Attention Transformer (FPAT) led to realistic and seamless foreground blending. Through various experiments and comparative analysis, we demonstrated that CHANGER offers significant qualitative and quantitative improvements over the state-of-the-art model. \textcolor{black}{We believe that the superiority and cost-effectiveness of CHANGER pave the way for its adoption in real-world applications, offering robust solutions for generating high-quality head blending contents.}

\clearpage

\section*{Acknowledgements}

This work was supported by the National Research Foundation of Korea (NRF) grant funded by the Korea government (MSIT) (No. 2710018251), and Korea Planning \& Evaluation Institute of Industrial Technology (KEIT) grant funded by the Korea government (MOTIE) (RS-2024-00444344), and in part by the IITP grant funded by the Korea Government (MSIT) (Artificial Intelligence Innovation Hub) under Grant 2021-0-02068, and by the IITP grant funded by the Korea government (MSIT) (No.RS-2022-00155911, Artificial Intelligence Convergence Innovation Human Resources Development (Kyung Hee University)).
{\small
\bibliographystyle{ieee_fullname}
\bibliography{main}

\begin{thebibliography}{10}\itemsep=-1pt

\bibitem{chen2020simswap}
Renwang Chen, Xuanhong Chen, Bingbing Ni, and Yanhao Ge.
\newblock Simswap: An efficient framework for high fidelity face swapping.
\newblock In {\em Proceedings of the 28th ACM International Conference on Multimedia}, pages 2003--2011, 2020.

\bibitem{Chung18b}
J.~S. Chung, A. Nagrani, and A. Zisserman.
\newblock Voxceleb2: Deep speaker recognition.
\newblock In {\em INTERSPEECH}, 2018.

\bibitem{dhariwal2021diffusion}
Prafulla Dhariwal and Alexander Nichol.
\newblock Diffusion models beat gans on image synthesis.
\newblock {\em Advances in Neural Information Processing Systems}, 34:8780--8794, 2021.

\bibitem{dosovitskiy2020image}
Alexey Dosovitskiy, Lucas Beyer, Alexander Kolesnikov, Dirk Weissenborn, Xiaohua Zhai, Thomas Unterthiner, Mostafa Dehghani, Matthias Minderer, Georg Heigold, Sylvain Gelly, et~al.
\newblock An image is worth 16x16 words: Transformers for image recognition at scale.
\newblock In {\em International Conference on Learning Representations}, 2020.

\bibitem{d2021convit}
St{\'e}phane d’Ascoli, Hugo Touvron, Matthew~L Leavitt, Ari~S Morcos, Giulio Biroli, and Levent Sagun.
\newblock Convit: Improving vision transformers with soft convolutional inductive biases.
\newblock In {\em International Conference on Machine Learning}, pages 2286--2296. PMLR, 2021.

\bibitem{kang2022oamixer}
Hyunwoo Kang, Sangwoo Mo, and Jinwoo Shin.
\newblock Oamixer: Object-aware mixing layer for vision transformers.
\newblock {\em arXiv preprint arXiv:2212.06595}, 2022.

\bibitem{kingma2014adam}
Diederik~P Kingma and Jimmy Ba.
\newblock Adam: A method for stochastic optimization.
\newblock {\em arXiv preprint arXiv:1412.6980}, 2014.

\bibitem{liu2023fine}
Zhian Liu, Maomao Li, Yong Zhang, Cairong Wang, Qi Zhang, Jue Wang, and Yongwei Nie.
\newblock Fine-grained face swapping via regional gan inversion.
\newblock In {\em Proceedings of the IEEE/CVF conference on computer vision and pattern recognition}, pages 8578--8587, 2023.

\bibitem{liu2021swin}
Ze Liu, Yutong Lin, Yue Cao, Han Hu, Yixuan Wei, Zheng Zhang, Stephen Lin, and Baining Guo.
\newblock Swin transformer: Hierarchical vision transformer using shifted windows.
\newblock In {\em Proceedings of the IEEE/CVF international conference on computer vision}, pages 10012--10022, 2021.

\bibitem{milletari2016v}
Fausto Milletari, Nassir Navab, and Seyed-Ahmad Ahmadi.
\newblock V-net: Fully convolutional neural networks for volumetric medical image segmentation.
\newblock In {\em 2016 fourth international conference on 3D vision (3DV)}, pages 565--571. Ieee, 2016.

\bibitem{Nagrani17}
A. Nagrani, J.~S. Chung, and A. Zisserman.
\newblock Voxceleb: a large-scale speaker identification dataset.
\newblock In {\em INTERSPEECH}, 2017.

\bibitem{quattoni2009recognizing}
Ariadna Quattoni and Antonio Torralba.
\newblock Recognizing indoor scenes.
\newblock In {\em 2009 IEEE conference on computer vision and pattern recognition}, pages 413--420. IEEE, 2009.

\bibitem{rombach2022high}
Robin Rombach, Andreas Blattmann, Dominik Lorenz, Patrick Esser, and Bj{\"o}rn Ommer.
\newblock High-resolution image synthesis with latent diffusion models.
\newblock In {\em Proceedings of the IEEE/CVF Conference on Computer Vision and Pattern Recognition}, pages 10684--10695, 2022.

\bibitem{shimoda1989new}
Shigeru Shimoda, Masaki Hayashi, and Yasuaki Kanatsugu.
\newblock New chroma-key imagining technique with hi-vision background.
\newblock {\em IEEE Transactions on broadcasting}, 35(4):357--361, 1989.

\bibitem{shiohara2023blendface}
Kaede Shiohara, Xingchao Yang, and Takafumi Taketomi.
\newblock Blendface: Re-designing identity encoders for face-swapping.
\newblock In {\em Proceedings of the IEEE/CVF International Conference on Computer Vision}, pages 7634--7644, 2023.

\bibitem{shu2022few}
Changyong Shu, Hemao Wu, Hang Zhou, Jiaming Liu, Zhibin Hong, Changxing Ding, Junyu Han, Jingtuo Liu, Errui Ding, and Jingdong Wang.
\newblock Few-shot head swapping in the wild.
\newblock In {\em Proceedings of the IEEE/CVF Conference on Computer Vision and Pattern Recognition}, pages 10789--10798, 2022.

\bibitem{simonyan2014very}
Karen Simonyan and Andrew Zisserman.
\newblock Very deep convolutional networks for large-scale image recognition.
\newblock {\em arXiv preprint arXiv:1409.1556}, 2014.

\bibitem{vaswani2017attention}
Ashish Vaswani, Noam Shazeer, Niki Parmar, Jakob Uszkoreit, Llion Jones, Aidan~N Gomez, {\L}ukasz Kaiser, and Illia Polosukhin.
\newblock Attention is all you need.
\newblock {\em Advances in neural information processing systems}, 30, 2017.

\bibitem{xu2022region}
Chao Xu, Jiangning Zhang, Miao Hua, Qian He, Zili Yi, and Yong Liu.
\newblock Region-aware face swapping.
\newblock In {\em Proceedings of the IEEE/CVF Conference on Computer Vision and Pattern Recognition}, pages 7632--7641, 2022.

\bibitem{yang2023paint}
Binxin Yang, Shuyang Gu, Bo Zhang, Ting Zhang, Xuejin Chen, Xiaoyan Sun, Dong Chen, and Fang Wen.
\newblock Paint by example: Exemplar-based image editing with diffusion models.
\newblock In {\em Proceedings of the IEEE/CVF Conference on Computer Vision and Pattern Recognition}, pages 18381--18391, 2023.

\bibitem{yoo2023fastswap}
Sahng-Min Yoo, Tae-Min Choi, Jae-Woo Choi, and Jong-Hwan Kim.
\newblock Fastswap: A lightweight one-stage framework for real-time face swapping.
\newblock In {\em Proceedings of the IEEE/CVF Winter Conference on Applications of Computer Vision}, pages 3558--3567, 2023.

\bibitem{yu2018bisenet}
Changqian Yu, Jingbo Wang, Chao Peng, Changxin Gao, Gang Yu, and Nong Sang.
\newblock Bisenet: Bilateral segmentation network for real-time semantic segmentation.
\newblock In {\em Proceedings of the European conference on computer vision (ECCV)}, pages 325--341, 2018.

\bibitem{zhang2020cross}
Pan Zhang, Bo Zhang, Dong Chen, Lu Yuan, and Fang Wen.
\newblock Cross-domain correspondence learning for exemplar-based image translation.
\newblock In {\em Proceedings of the IEEE/CVF Conference on Computer Vision and Pattern Recognition}, pages 5143--5153, 2020.

\bibitem{zhang2021flow}
Zhimeng Zhang, Lincheng Li, Yu Ding, and Changjie Fan.
\newblock Flow-guided one-shot talking face generation with a high-resolution audio-visual dataset.
\newblock In {\em Proceedings of the IEEE/CVF Conference on Computer Vision and Pattern Recognition}, pages 3661--3670, 2021.

\bibitem{zhou2023propainter}
Shangchen Zhou, Chongyi Li, Kelvin~CK Chan, and Chen~Change Loy.
\newblock Propainter: Improving propagation and transformer for video inpainting.
\newblock In {\em Proceedings of the IEEE/CVF International Conference on Computer Vision}, pages 10477--10486, 2023.

\end{thebibliography}
}

\clearpage
\setcounter{page}{1}
\setcounter{figure}{0}
\setcounter{table}{0}
\setcounter{section}{0}

\section*{Supplementary Material}


\begin{figure*}[t]
\begin{center}
\includegraphics[width=2.1\columnwidth]{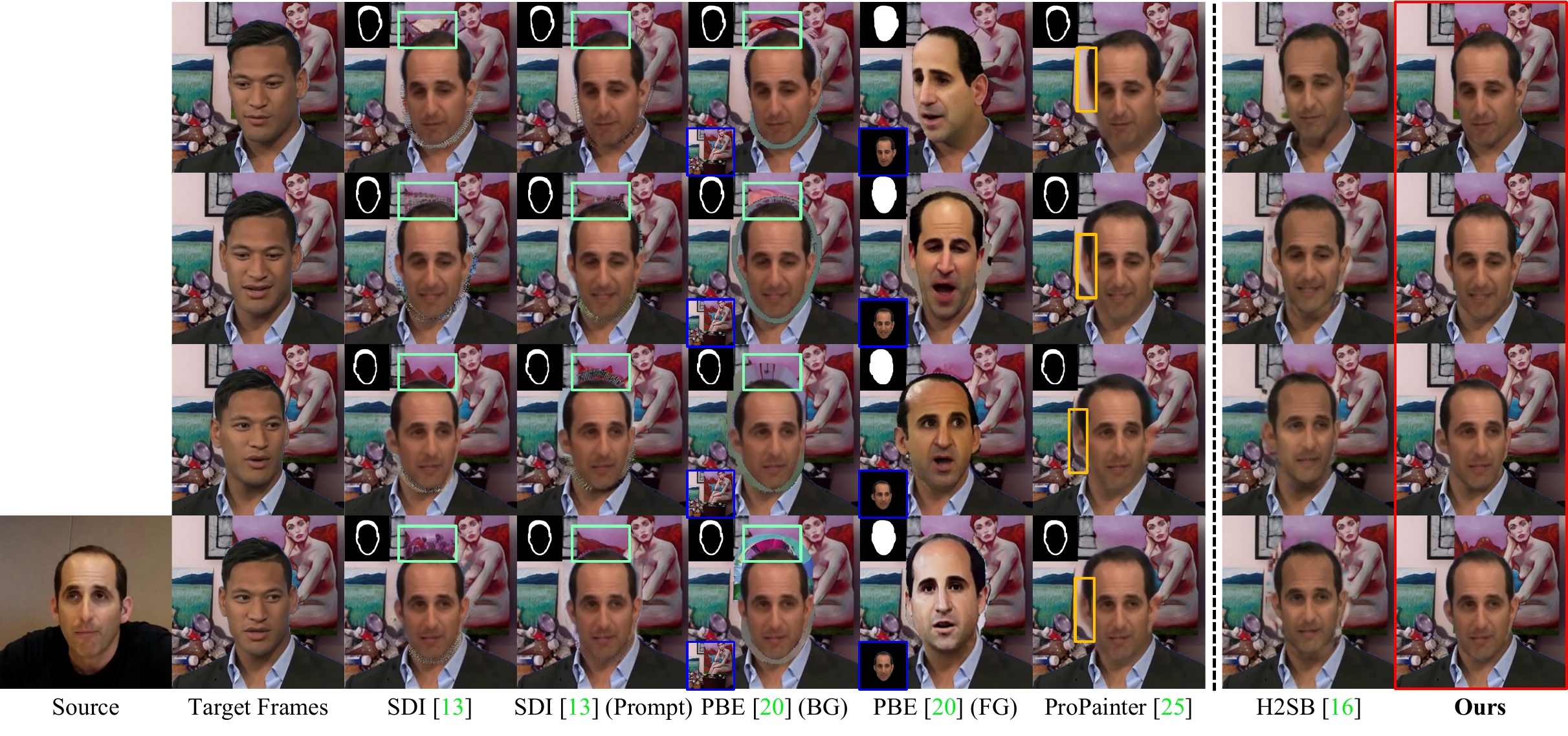}
\end{center}
\vspace{-5mm}
\caption{
Qualitative comparisons of using recent inpainting baselines~\cite{rombach2022high, yang2023paint, zhou2023propainter} and the head blending model~\cite{shu2022few} on sequential frames of a target video. We tested both scenarios with and without text prompting (Prompt) for \textit{SDI}.
For \textit{PBE}, we separated scenarios; the background (BG) and the foreground (FG) references (bottom-left blue boxes of each column).
 }
\label{fig:inpaint}
\end{figure*}

This supplementary document provides an intensive insight into our work presented in the main paper, consisting of qualitative comparisons with the state-of-the-art inpainting methods, the notation and visualization of our Head shape and long Hair ($H^2$) augmentation, more detailed analysis and descriptions of the proposed Foreground Predictive Attention Transformer (FPAT), implementation details on the training objectives, experimental details on the user study, and a head blending video results in our project page.

\renewcommand{\thesection}{\Alph{section}}

\section{Qualitative Comparisons with Recent Inpainting Models}

In this section, we investigate the performance of our CHANGER compared to the state-of-the-art inpainting models through qualitative comparisons.

\noindent\textbf{Baselines.} We establish the state-of-the-art inpainting models as follows: (1) Stable Diffusion Inpainting (SDI)~\cite{rombach2022high}, (2) Paint-by-Example (PBE)~\cite{yang2023paint}, (3) ProPainter~\cite{zhou2023propainter}.

Figure~\ref{fig:inpaint} shows the results from head blending video compared with the recent diffusion-based or video-based inpainting models. \textit{SDI} and \textit{PBE} mainly suffered from background generation (green boxes) and artifacts of the foreground region. 
\textit{ProPainter} showed blurry foreground generation (orange boxes). 
Our results show not only the highest fidelity in the background inpainting region as well as the foreground but also stability in a time-consistency perspective, which ensures the quality of the video output.

\begin{table*}[t]
\small
\centering
\begin{tabular}{ccc}
\toprule
\textbf{Notation} & \textbf{Dimension} & \textbf{Description} \\ \midrule
$I_S$ & $\mathbb{R}^{ 3 \times H \times W}$ & Source image.\\ 
$I_T$ & $\mathbb{R}^{ 3 \times H \times W}$ & Target image.\\ 
$I^{gray}_{S}$ & $\mathbb{R}^{ 1 \times H \times W}$ & Gray-scale image from source.\\ 
$I_T^{green}$ & $\mathbb{R}^{ 3 \times H \times W}$ & Target image with a green screen background.\\ 
\multirow{2}{*}{$I^{head}_T$} & \multirow{2}{*}{$\mathbb{R}^{ 3 \times H \times W}$} & Target head image, used in Head Colorizer, \\
& & made by only leaving the head region from the target image.\\
\multirow{2}{*}{$I^{body}_T$} & \multirow{2}{*}{$\mathbb{R}^{ 3 \times H \times W}$} & Target body image, used in Body Blender, \\
& &  made by excluding head, neck, and background.\\
$\mathtt{M}^{head}_{S}$ & $\mathbb{R}^{ 1 \times H \times W}$ & Head mask from source.\\
$\mathtt{M}^{head}_{T}$ & $\mathbb{R}^{ 1 \times H \times W}$ & Head mask from target.\\
$\mathtt{M}^{head}_{h^1}$ & $\mathbb{R}^{ 1 \times H \times W}$ & Augmented head mask made by transformation $\mathcal{T}_{head}$. \\
\multirow{2}{*}{$\mathtt{M}^{head}_{h^2}$} & \multirow{2}{*}{$\mathbb{R}^{ 1 \times H \times W}$} & Augmented head mask from $\mathtt{M}^{head}_{h^1}$, \\
& & made by transformation $\mathcal{T}_{hair}$. \\
\multirow{2}{*}{$\mathtt{M}^{head}_{union}$} & \multirow{2}{*}{$\mathbb{R}^{ 1 \times H \times W}$} & Union mask of $\mathtt{M}^{head}_{h^2}$ and $\mathtt{M}^{head}_{S}$ during training, \\
& & union mask of $\mathtt{M}^{head}_{T}$ and $\mathtt{M}^{head}_{S}$ during testing.\\

$\mathtt{M}^{ip}$ & $\mathbb{R}^{ 1 \times H \times W}$ & Inpainting region subtracting $\mathtt{M}^{head}_{S}$ from $\mathtt{M}^{head}_{union}$.\\

\multirow{2}{*}{$M$} & \multirow{2}{*}{$\mathbb{R}^{ 1 \times H \times W}$} & Predicted foreground mask which is further used \\
& & as an input of the FPAT blocks.\\
$X$ & $\mathbb{R}^{ 3 \times H \times W}$ & Input for our CHANGER.\\
$Y$ & $\mathbb{R}^{ 3 \times H \times W}$ & Head blended outputs of our CHANGER.\\\bottomrule
\end{tabular}
\caption{\textbf{Notations and corresponding descriptions in our CHANGER.}
}
\label{tab:suppnotation}
\end{table*}
\begin{figure*}[t]
\begin{center}
\includegraphics[width=2.1\columnwidth]{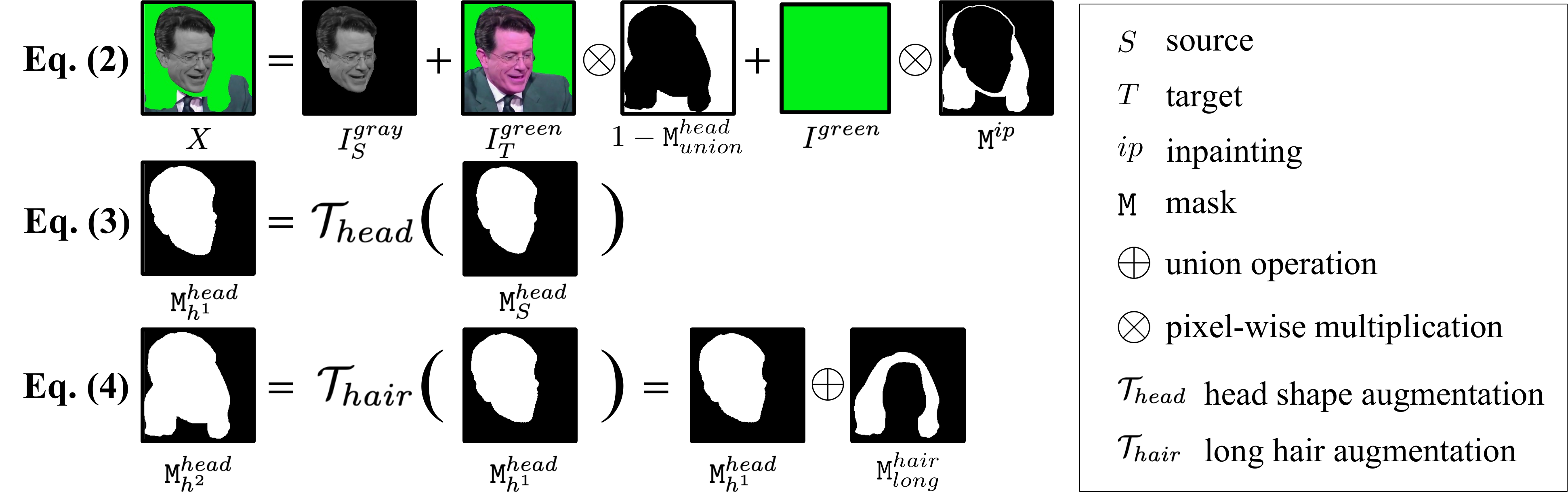}
\end{center}
\caption{
\textbf{Visualization of $\bm{H^2}$ Augmentation.} 
Eq. (2) is the input $X$ formulation during training. Inspired by~\cite{yoo2023fastswap}, we apply the same color jitter to both $I_{T}^{green}$ and the ground truth during the training phase. 
Eq. (3) shows the head shape augmentation.
Eq. (4) shows the long hair augmentation.
}
\label{fig:equation}
\end{figure*}

\section{Notation and Visualization Summary of \boldmath $H^2$ Augmentation}
We provide a detailed explanation of the various notations used in our method, especially for the proposed Head shape and long Hair ($H^2$) augmentation, in Table \ref{tab:suppnotation}. We also visualize the process of $H^2$ augmentation in Figure \ref{fig:equation}. Please refer to the descriptions in the table and figure, to ensure clarity in interpreting our work.

\begin{figure*}[t]
\begin{center}
\includegraphics[width=2.1\columnwidth]{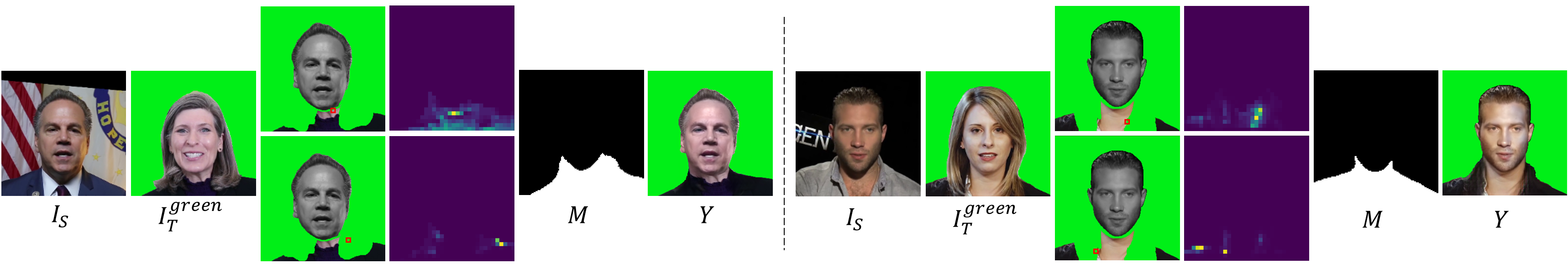}
\end{center}
\caption{
The foreground mask predicted by FPAT (\textit{$M$}), the attention map used in the transformer layer (\textit{Attention}), and the head blending result (\textit{$Y$}) when input source image $I_S$ and target image $I_T$ are used. We visualize the similarity between the query patch (red box) and each key patch in the depicted image as an attention map. Blue represents low values and yellow represents high values.}
\label{fig:supp_attention}
\end{figure*}

\section{More Details in FPAT}
\subsection{Attention Map of FPAT}
The Foreground Predictive Attention Transformer (FPAT) stands at the forefront of our model structure, primarily focusing on the enhancement of the fidelity of foreground blending. In the diverse situations created by various head shape and hairstyle differences between the source and target images, FPAT aims to predict the foreground region and then attend to the predicted foreground region. We demonstrate the effectiveness of the FPAT qualitatively. 

Figure \ref{fig:supp_attention} presents predicted masks ($M$), attention maps (\textit{Attention}), and head blending results ($Y$) obtained by FPAT on various source and target pairs. For each input, two distinct attention maps are depicted: one for the neck (upper row) and another for the cloth (lower row). The small red boxes inside the images in the $X$ column represent the patches used to generate queries for our proposed FPAT transformer layer. The images in the \textit{Attention} column depict the calculated attention derived from these queries and keys, where higher values are represented closer to yellow and lower values closer to blue.

The predicted mask results show that FPAT effectively reconstructs obscured foreground areas caused by long hair. Furthermore, meaningful attention is trained within the predicted region, as depicted in the attention maps. Specifically, during the generation of the neck region (upper row), the model focuses explicitly on the neck area of the target image. In contrast, when generating the occluded attire region (lower row), the model focuses on relevant clothing areas, indicating its ability to create images with attention to pertinent regions.

\subsection{Detailed Explanation of FPAT Mechanism}

Our FPAT starts with the input $z_c$, and predicts a foreground region, including the body and the neck, as a binary mask $ M \in \mathbb{R}^{h \times w} $ with Foreground-Prediction module. The FPAT block refers to the target body information $I_T^{body}$ and updates $z_c$ using the information of $M$ to generate the neck and body via the Foreground-Aware Transformer block.
FPAT patchifies the feature output of the head colorizer $z_c \in \mathbb{R}^{C \times h \times w}$ and get $z_c^p \in \mathbb{R}^{N \times P^2C}$, where $(P, P)$ is the resolution of the patches and $N=hw/P^2$ is the number of patches. FPAT also patchifies the embedded feature of the target body $I_{T}^{body}$ as $z_{body}^p \in \mathbb{R}^{N \times P^2C}$, and the predicted body and neck mask $M$ as $M^p \in \mathbb{R}^{N \times P^2}$. Then, FPAT averages $M^p$ along the channel axis to acquire $M^p_{\texttt{avg}} \in \mathbb{R}^{N}$ as following:
\begin{equation}
    [M^p_{\texttt{avg}}]_n = \frac{1}{P^2} \Sigma_{m=1}^{P^2} M^p_{nm},
\end{equation}
where $[M^p_{\texttt{avg}}]_n$ is the $n$-th patch of $M^p_{\texttt{avg}}$ and $M^p_{nm}$ is the ${(n, m)}$-th element of $M^p$. Next, we divide patches into two groups: (1) a set of patches $S_{b}$ that includes the predicted body and neck parts and (2) a set of patches $S_{nb}$ that does not include them by thresholding $M^p_{avg}$ by following:
\begin{equation}
\begin{split}
    \label{eq:tau}
    S_{b} = \{ i \in 1, ..., N \,|\, [M^p_{\texttt{avg}}]_i \geq \tau \} \\
    S_{nb} = \{ i \in 1, ..., N \,|\, [M^p_{\texttt{avg}}]_i < \tau \},
\end{split}
\end{equation}
where $\tau$ is the hyperparameter.
Then, FPAT computes the binary mask $M^b \in \mathbb{R}^{N \times N}$ as following:
\begin{equation}
    M_{ij}^b=\begin{cases}
    0, & \text{if } \: i, j \in S_{b} \: \text{ , } \: i, j \in S_{nb},\\
    -\infty, & \text{otherwise},
  \end{cases}
\end{equation}
where $M_{ij}^b$ is the $(i, j)$-th element of $M^b$.

Finally, FPAT masks the attention between a query from the latent representation $z_c^p$, key and value from the target head feature $z_{body}^p$.

\section{Training Objectives Details}
We train the model with $\mathcal{L}_{total}$, which is a summation of \textbf{(1)} $\mathcal{L}_{rec}$, the reconstruction loss for the final output head and the ground truth, \textbf{(2)} $\mathcal{L}_{hc}$, the reconstruction loss for the output of $\texttt{ToRGB}$ block, \textbf{(3)} $\mathcal{L}_{mask}$~\cite{milletari2016v}, the loss for the output of the Foreground-Prediction module, \textbf{(4)} perceptual loss $\mathcal{L}_{per}$, and \textbf{(5)} adversarial loss $\mathcal{L}_{adv}$ for our objective functions.

Corresponding objective functions are as follows:

\begin{align}
\begin{split}\label{eq:12}
    \mathcal{L}_{total} &= \lambda_{rec} \mathcal{L}_{rec} + \lambda_{hc}\mathcal{L}_{hc} + \lambda_{mask}\mathcal{L}_{mask} \\ &+ \lambda_{per} \mathcal{L}_{per} + \lambda_{adv} \mathcal{L}_{adv},
\end{split}\\
    \mathcal{L}_{rec} &= \ ||Y \otimes \mathtt{M}^{head}_{S} - I_T \otimes \mathtt{M}^{head}_{S}||_1,
\\
    \mathcal{L}_{hc} &= \ ||Y - I^{hc}_{T}||_1,
\\
    \mathcal{L}_{mask} &= \ ||M_{gt} - M||_1,
\\
    \mathcal{L}_{per} &= \ \sum^{L}_{i=1}||\Phi_i(Y) - \Phi_i(I_T)||_1,
\\
\begin{split}\label{eq:16}
    \mathcal{L}^{D_I}_{adv} &= - \mathbb{E}_{I_T\sim p_{data}}[\text{log}(D_I(I_T))] \\ 
    &- \mathbb{E}_{Y\sim p_Y}[\text{log}(1-D_I(Y))] ,
\end{split}\\
    \mathcal{L}^{\mathcal{D}(z)}_{adv} &= - \mathbb{E}_{Y\sim p_Y}[D_I(Y)],
\end{align}
where $\lambda_{rec}$, $\lambda_{hc}$, $\lambda_{mask}$, $\lambda_{per}$, and $\lambda_{adv}$ are weights for the loss $\mathcal{L}_{rec}$, $\mathcal{L}_{hc}$, $\mathcal{L}_{mask}$, $\mathcal{L}_{per}$, and $\mathcal{L}_{adv}$, respectively. $I^{hc}_{T}$ is a target image without neck, and body completion and $\Phi$ is a pre-trained VGG19 network~\cite{simonyan2014very}, and $D_I$ is a discriminator.
We used $\lambda_{rec}$ = 10, $\lambda_{hc}$ = 10, $\lambda_{mask}$ = 10, $\lambda_{per}$ = 1, and $\lambda_{adv}$ = 1.

\section{User Study}
We elucidate the details of our user study in the attached \textit{User Study.pdf}. The material includes the questionnaire design, participant demographics, and methodology. We also present the comprehensive results in Excel format.

\section{Project Page}
The head blending video results are shown on our project page linked in the footnote of the main paper. The video results demonstrate the effectiveness and robustness of CHANGER in various industrial scenarios and suggest its potential for adoption in the industrial field. We submit a project page created in HTML to firmly prove the finality and completeness of the results displayed on our project page as of the submission date.



\end{document}